\documentclass[preprint]{article}

\usepackage{neurips_2023}
\usepackage[authoryear]{natbib}

\usepackage{bookmark}
\usepackage[utf8]{inputenc} 
\usepackage[T1]{fontenc}
\usepackage{hyperref} 
\usepackage{url}
\usepackage{booktabs} 
\usepackage{amsfonts} 
\usepackage{nicefrac} 
\usepackage{microtype}
\usepackage{xcolor} 
\usepackage{listings}
\usepackage{cite}
\usepackage{graphicx}
\usepackage{hyperref}
\usepackage{amsmath}

\makeatletter
\@addtoreset{equation}{section}
\makeatother
\renewenvironment{equation}{\begin{equation*}}{\end{equation*}}
\usepackage{algpseudocode}
\usepackage{algorithm}
\usepackage{caption}
\usepackage{pdflscape}
\usepackage{tabularx}
\usepackage{multirow}
\usepackage{booktabs}
\usepackage{float}
\usepackage{adjustbox}

\hypersetup{
pdftitle={Improving X-Codec-2.0 for Multi-Lingual Speech: 25 Hz Latent Rate and 24 kHz Sampling},
pdfauthor={Husein Zolkepli},
pdfsubject={Audio and Speech Processing},
pdfkeywords={speech models, deep learning},
colorlinks=true,
linkcolor=blue,
citecolor=blue,
urlcolor=blue
}
\title{Improving X-Codec-2.0 for Multi-Lingual Speech: 25 Hz Latent Rate and 24 kHz Sampling}
\author{
  Husein Zolkepli \\
  Scicom (MSC) Berhad, Malaysia \\
  \texttt{husein.zolkepli@scicom.com.my}
}

\begin{document}

\maketitle

\begin{abstract}

X-Codec-2.0 has shown strong performance in neural audio compression and multilingual speech modeling, operating at a 50 Hz latent rate and a 16 kHz sampling rate using frozen HuBERT features. While effective, this configuration limits temporal efficiency and audio fidelity.
In this work, we explore a simple and effective modification by introducing additional pooling and increasing the decoder hop size. This reduces the latent rate from 50 Hz to 25 Hz and simultaneously raises the output sampling rate from 16 kHz to 24 kHz, improving efficiency and perceptual quality without altering the core architecture.
Evaluated on the multilingual Common Voice 17 test set, the proposed configuration achieves a 0.29 MOS improvement over the original X-Codec-2.0 baseline based on UTMOSv2, and attains the best reported performance among all codecs operating at 25 Hz. The source code, checkpoints, and generation comparisons are released at \href{https://huggingface.co/Scicom-intl/xcodec2-25TPS-24k}{https://huggingface.co/Scicom-intl/xcodec2-25TPS-24k}.

\end{abstract}

\section{Introduction}

Recent advances in neural audio tokenization have enabled end-to-end speech modeling through discrete latent representations.  
Among these, X-Codec-2.0 (\citet{ye2024codecdoesmatterexploring}) has emerged as a powerful multilingual codec model, combining a HuBERT-based (\citet{hsu2021hubertselfsupervisedspeechrepresentation}) semantic encoder and a transformer-based codec architecture to produce high-quality, language-agnostic audio tokens.  
Its design allows large language models (LLMs) and autoregressive decoders to handle speech as a discrete sequence prediction task.  

A key appeal of X-Codec-2.0 lies in its simplicity, it employs a single codebook with a latent frame rate of 50 Hz, making its discrete token stream compact and easy to integrate into multimodal LLM pipelines.
This property makes it particularly attractive for tasks such as text-to-speech (TTS) generation, speech-to-text modeling, or unified speech-language pretraining, especially in streaming or low-latency scenarios where token-level autoregressive decoding is desirable.
Furthermore, the use of fused linear cross-entropy in modern LLM implementations can substantially reduce activation memory during training, enabling efficient extension of large language models to handle audio tokens without prohibitive computational overhead.

However, X-Codec-2.0 operates at a 50\,Hz latent frame rate with a 16\,kHz target sampling rate.  
While effective, this configuration limits temporal resolution and upper-frequency fidelity, producing slightly muffled high-frequency content and longer token sequences for generation models.  
Moreover, as multilingual datasets expand to include more diverse acoustic and expressive conditions, the fixed 50\,Hz resolution may underutilize the model's potential to capture fine-grained speech variation.

In this work, we propose a simple modification to X-Codec-2.0 that improves both efficiency and perceptual quality.  
By increasing the hop size to 960 samples and introducing a lightweight pooling layer before quantization, the codec operates at a reduced 25\,Hz latent rate while simultaneously increasing the audio sampling rate to 24\,kHz.  
All encoder components are kept frozen, and only the decoder is fine-tuned under this new temporal configuration.  
We find that this adjustment produces higher-quality reconstructions without additional parameters or training complexity.

Empirically, the proposed model achieves a +0.29 improvement in mean opinion score (MOS) as estimated by UTMOSv2 (\citet{baba2024t05voicemoschallenge2024}) on the multilingual Common Voice (\citet{ardila2020commonvoicemassivelymultilingualspeech}) 17 test set.  
The improvement is consistent across languages and demonstrates better high-frequency reconstruction and overall perceptual clarity.  
Our work shows that small architectural refinements by changing hop-size and pooling adjustments can meaningfully improve codec quality while preserving the simplicity and modularity that make X-Codec-2.0 suitable for LLM-based speech modeling.

\section{Method}
\label{sec:method}

The proposed modification to X-Codec-2.0 focuses on improving temporal efficiency and output audio fidelity through minimal architectural changes. The overall structure remains consistent with the original design, consisting of a frozen semantic encoder (HuBERT-based), a transformer codec encoder, and a vocoder-style decoder.

\subsection{Temporal Pooling and Hop Size Adjustment}
In the original X-Codec-2.0 configuration, the encoder operates at a latent rate of 50\,Hz with a 16\,kHz sampling rate, corresponding to a hop size of 320 samples. To achieve a lower latent rate and higher waveform sampling rate, we increased the hop size to 960 samples and introduced an additional average pooling layer:
\[
\text{AvgPool1d}(k=2,\, \text{stride}=2).
\]
This modification reduces the latent rate from 50\,Hz to 25\,Hz, halving the number of discrete tokens per second while maintaining temporal coherence. The pooling operation is applied before vector quantization, compressing the feature sequence by a factor of two.

\subsection{Decoder Weight Interpolation}
Changing the hop size also alters the dimensionality of the decoder's output layer. Rather than discarding the pretrained decoder weights, we applied one-dimensional linear interpolation to the output projection parameters of the generator head:
\begin{equation}
w'_i = (1 - \alpha_i)\, w_{\lfloor x_i \rfloor} + \alpha_i\, w_{\lceil x_i \rceil}, 
\quad 
x_i = \frac{L - 1}{L' - 1} \, i,
\quad 
\alpha_i = x_i - \lfloor x_i \rfloor,
\quad 
i = 0, \ldots, L' - 1
\end{equation}
Both the output weight and bias were interpolated to match the new hop size (960 samples). 
This procedure allows the decoder to retain the spectral characteristics of the original pretrained model while adapting smoothly to the new resolution. 
Although this interpolation strategy has not been empirically validated in isolation, it provided a straightforward way to transfer pretrained weights without retraining from scratch. 
We did not conduct ablation experiments with randomly initialized parameters, so it remains unclear whether interpolation contributes to faster convergence or improved stability during fine-tuning.

\subsection{Parameter Freezing and Adaptation}
All model parameters were frozen except for the decoder. The semantic encoder (frozen HuBERT) and codec encoder were reused directly from the pretrained X-Codec-2.0 checkpoint. The decoder was fine-tuned to accommodate the new hop size and temporal pooling. The resulting model generates 25\,Hz discrete tokens and reconstructs 24\,kHz audio with improved perceptual quality.

\section{Experiments}

\subsection{Training Datasets}
We train our model on a large-scale multilingual corpus totaling approximately 16,000 hours of speech. 
The dataset combines publicly available TTS and expressive speech corpora from Hugging Face, covering over 100 languages, including English, Mandarin, Malay, Japanese, Korean, Arabic (with various regional dialects), Hindi, Tamil, Bengali, and a wide range of Indic, European, and other low-resource languages.
Each audio sample is uniformly resampled to 24\,kHz and cropped into randomly selected 5-second segments. 
No text transcriptions are used during training. The full list of datasets is publicly available at Malaysia-AI Multilingual-TTS repository. \footnote{\url{https://huggingface.co/datasets/malaysia-ai/Multilingual-TTS/blob/2421a13e07226d96ac7009d5327d96a84672768c/README.md}}

\subsection{Model Initialization}
We initialize from the official X-Codec-2.0 checkpoint provided by HKUSTAudio.
As described in Section~\ref{sec:method}, the semantic encoder and codec encoder are frozen, while only the decoder is fine-tuned under the modified hop size (960 samples) and 25\,Hz latent rate.
Decoder head weights and biases are linearly interpolated from the pretrained model to match the new output dimensionality.

\subsection{Training Configuration}
All hyperparameters were kept consistent with the original X-Codec-2.0 implementation to ensure comparability. The model was trained on two NVIDIA RTX 3090 Ti GPUs using BF16 mixed precision for 3 million steps with a batch size of 20 per device. Both the generator and discriminator were optimized using the Adam optimizer (\(\beta_1 = 0.8, \beta_2 = 0.9\)) and a cyclic learning rate schedule with warmup and decay:
\[
\text{max\_lr} = 1.0\times10^{-4}, \quad
\text{min\_lr} = 2.0\times10^{-5}
\]
The loss function followed the same multi-objective formulation as X-Codec-2.0, combining mel-spectrogram, adversarial, and semantic losses except for feature matching:
\[
\mathcal{L}_{\text{total}} =
\lambda_{\text{mel}}\mathcal{L}_{\text{mel}} +
\lambda_{\text{adv}}\mathcal{L}_{\text{adv}} +
\lambda_{\text{sem}}\mathcal{L}_{\text{sem}}
\]
Empirically, the following coefficients were retained from the original configuration:
\(\lambda_{\text{mel}}=15\),
\(\lambda_{\text{adv}}=1\),
\(\lambda_{\text{sem}}=5\).

To align with the new sampling rate, the \texttt{MultiResolutionMelSpectrogramLoss} was configured for a 24\,kHz sample rate. This ensures that spectral loss computation matches the decoder's output frequency resolution. Validation was performed every 4000 steps on 2560 held-out samples. Gradient clipping was set to 1.0 for both generator and discriminator to maintain stability. The overall setup reproduces the X-Codec-2.0 training dynamics while isolating the effect of the proposed hop-size and pooling modifications.

\section{Evaluation}
We evaluate perceptual quality using the UTMOSv2, a neural predictor trained to approximate human mean opinion scores (MOS) from speech spectrograms. 
Although UTMOSv2 is trained primarily on English speech, it has demonstrated strong correlation with subjective evaluations across diverse acoustic conditions and can be reasonably applied to multilingual speech quality assessment.

\subsection{Evaluation Dataset}
We use the multilingual Common Voice~17 test set, which covers 116 languages. 
All audio samples are preprocessed using the WebRTC Voice Activity Detection (VAD) to remove non-speech regions, including both leading, trailing, and intermediate silences, with a minimum silence duration threshold of 0.2 seconds. 
For each language, utterances shorter than 20 seconds after trimming are retained, then sorted in descending order of duration. 
The 500 longest samples per language are selected for evaluation, resulting in a total of 48{,}489 audio clips.
The complete evaluation set is publicly available at Scicom-intl xcodec2-25TPS-24k repository. \footnote{\url{https://huggingface.co/Scicom-intl/xcodec2-25TPS-24k/tree/main/test-set}}

\subsection{Comparison with Other Models}
We further benchmark our model against several recent neural audio codecs: DAC (\citet{kumar2023highfidelityaudiocompressionimproved}), DistilCodec (\citet{wang2025unittsendtoendttsdecoupling}), Encodec (\citet{defossez2022highfidelityneuralaudio}), LongCat-Audio-Codec (\citet{zhao2025longcataudiocodecaudiotokenizerdetokenizer}), Mimi (\citet{defossez2024moshispeechtextfoundationmodel}), Neucodec (\citet{julian2025finitescalarquantizationenables}), Qwen3-TTS-Tokenizer (\citet{hu2026qwen3ttstechnicalreport}), SNAC (\citet{siuzdak2024snac}), SpeechTokenizer (\citet{zhang2024speechtokenizerunifiedspeechtokenizer}), UniCodec (\citet{jiang2025unicodecunifiedaudiocodec}), WavTokenizer (\citet{ji2025wavtokenizerefficientacousticdiscrete}) and X-Codec-2.0 baseline.

\begin{table*}[!htbp]
\centering
\caption{UTMOSv2 evaluation across representative languages from Common Voice~17.}
\label{tab:utmos-comparison}
\setlength{\tabcolsep}{6pt}
\scriptsize
\begin{tabular}{llcccccccccc}
\toprule
\textbf{Model} & \textbf{Codebook} & \textbf{Nq} & \textbf{Rate (Hz)} & \textbf{NL} & \textbf{EN} & \textbf{FR} & \textbf{IT} & \textbf{PL} & \textbf{PT} & \textbf{ES} \\
\midrule
DAC & 1024 & 9 & 86 & 2.315 & 2.037 & 2.085 & 2.200 & 2.258 & 2.231 & 2.204 \\
DistilCodec & 32768 & 1 & 93 & 2.416 & 2.130 & 2.141 & 2.261 & 2.312 & 2.308 & 2.194 \\
Encodec & 1024 & 2 & 75 & 1.388 & 1.464 & 1.386 & 1.414 & 1.424 & 1.405 & 1.369 \\
LongCat & 8192 & 4 & 16.6 & 2.500 & 2.341 & 2.226 & 2.364 & 2.424 & 2.396 & 2.326 \\
Mimi & 2048 & 32 & 12.5 & 2.336 & 2.032 & 2.081 & 2.152 & 2.296 & 2.273 & 2.201 \\
\textbf{Neucodec} & \textbf{65536} & \textbf{1} & \textbf{50} & \textbf{2.550} & \textbf{2.361} & \textbf{2.354} & \textbf{2.494} & \textbf{2.493} & \textbf{2.491} & \textbf{2.504} \\
Qwen3-TTS-Tokenizer & 4096 & 16 & 12 & 2.462 & 2.236 & 2.273 & 2.349 & 2.405 & 2.355 & 2.346 \\
SNAC & 4096 & 3 & 27 & 2.193 & 2.048 & 2.007 & 2.098 & 2.144 & 2.187 & 2.117 \\
SpeechTokenizer & 1024 & 8 & 50 & 2.203 & 2.065 & 2.003 & 2.062 & 2.159 & 2.202 & 2.096 \\
UniCodec & 16384 & 1 & 75 & 2.256 & 2.014 & 2.026 & 2.149 & 2.174 & 2.221 & 2.111 \\
WavTokenizer 40TPS & 4096 & 1 & 40 & 2.319 & 2.150 & 2.107 & 2.189 & 2.230 & 2.241 & 2.231 \\
WavTokenizer 75TPS & 4096 & 1 & 75 & 2.209 & 2.087 & 2.060 & 2.201 & 2.186 & 2.223 & 2.202 \\
X-Codec-2.0 (baseline) & 65536 & 1 & 50 & 2.168 & 2.086 & 2.062 & 2.108 & 2.122 & 2.136 & 2.139 \\
\midrule
\textbf{Ours (25\,Hz, 24\,kHz)} & \textbf{65536} & \textbf{1} & \textbf{25} & \textbf{2.457} & \textbf{2.245} & \textbf{2.217} & \textbf{2.304} & \textbf{2.392} & \textbf{2.386} & \textbf{2.314} \\
\bottomrule
\end{tabular}
\end{table*}

We report UTMOSv2 results on eight representative languages, Dutch, English, French, German, Italian, Polish, Portuguese, and Spanish, which collectively cover a range of phonetic and prosodic diversity among high-resource European languages. In addition to these representative results, we conduct UTMOSv2 evaluation on a total of 116 languages, with complete results reported in Scicom-intl xcodec2-25TPS-24k repository. \footnote{\url{https://github.com/Scicom-AI-Enterprise-Organization/X-Codec-2.0-25TPS-24k/tree/main/evaluation}}

\begin{figure}[h]
  \centering
  \includegraphics[width=0.8\linewidth]{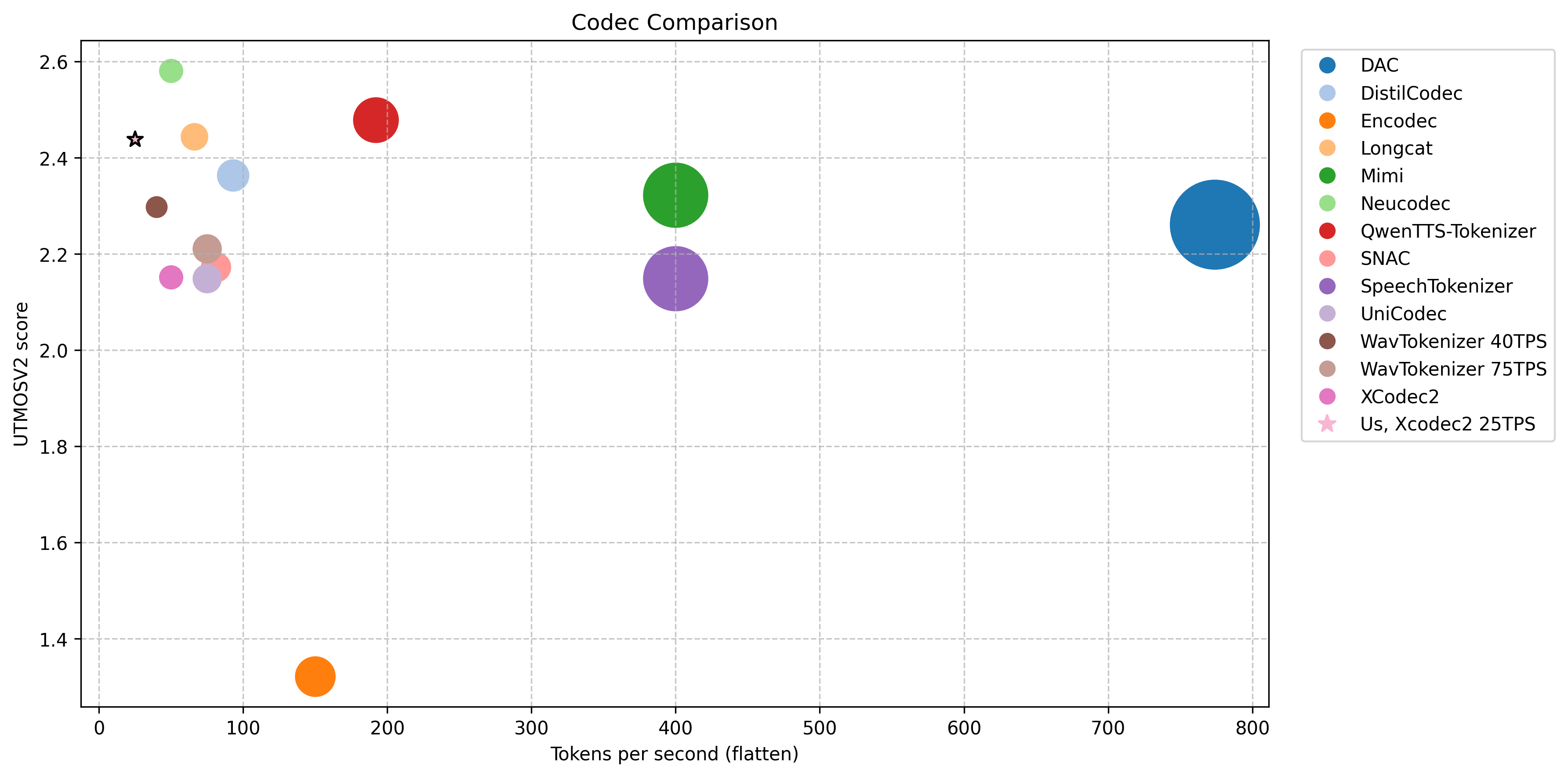}
  \caption{116 languages codec comparison}
\end{figure}

Figure above shows the aggregated results across all 116 languages in Common Voice 17. Under the fixed 25 Hz token rate constraint, our method consistently outperforms all competing codecs, establishing a new state of the art in the low token rate regime.

\section{Limitation}

While the proposed modification improves the perceptual quality and efficiency of X-Codec-2.0, several limitations remain.
First, the training data used in this study is primarily drawn from the Common Voice dataset, which is relatively clean and contains limited variation in background noise, speaker style, and expressiveness. As a result, the model does not generalize well to unseen languages or expressive speech domains such as animated or emotional voices. Continued fine-tuning with more diverse and expressive data is expected to alleviate this issue.

Second, all evaluations were conducted using the UTMOSv2 metric, which predicts mean opinion scores (MOS) directly from audio representations.
While this approach enables scalable and reproducible evaluation without the need for human raters, UTMOSv2 may not fully reflect subjective perceptual preferences.
Moreover, as the model was primarily trained on English speech, its cross-lingual generalization remains uncertain.
Although prior work has shown that UTMOSv2 maintains strong correlation with human judgments under diverse acoustic conditions, further validation on multilingual data is necessary to ensure fair and reliable assessment across languages.

Finally, no downstream applications have been explored. With a vocabulary size of 65,536 and a 25 Hz token rate, each discrete token represents a larger amount of information. This increases the effective prediction difficulty for autoregressive models that use these tokens, which may lead to higher perplexity.

\section{Future Work}

While this work focuses on a simple temporal modification introducing additional pooling and increasing the encoder hop size to reduce the latent rate from 50 Hz to 25 Hz in X-Codec-2.0 several directions remain to deepen the understanding and extend the capability of discrete audio representation models.

First, the semantic encoder in X-Codec-2.0, derived from a frozen HuBERT model, is known to capture cross-lingual acoustic and phonetic information effectively. However, its interaction with temporal resolution remains underexplored. Future work could systematically vary the latent frame rate (e.g., from 10 Hz to 100 Hz) and evaluate its impact on perceptual quality, token predictability, and downstream generative modeling. Such experiments would help characterize the decoder's temporal sensitivity and identify the optimal trade-off between information density and reconstruction quality. In particular, understanding how much semantic content is preserved when the latent rate decreases could inform the design of more compact and efficient tokenizers for both speech and general audio.

Second, reducing the latent rate introduces a natural compression bottleneck, potentially limiting the decoder's ability to recover fine-grained acoustic details. This raises an important design question regarding the balance between encoder compression strength and decoder capacity. Increasing the decoder depth, widening its receptive field, or incorporating attention-based upsampling mechanisms may help compensate for lost temporal detail when operating at lower frame rates. A systematic analysis of this trade-off to evaluate how decoder complexity scales with latent rate could provide new insights into optimal model scaling laws for discrete audio codecs.

Third, extending the experiments to larger and more diverse datasets, including noisy or expressive speech (e.g., emotional, conversational, or singing voice data), would reveal the model's robustness and generalization ability. The current results, based on primarily reflect performance on clean and monotonic speech, future work could investigate whether adjusting the latent frame rate interacts differently with noisy or highly variable input conditions.

Finally, downstream evaluation remains an open area. One promising direction is to assess how the discrete tokens perform in text-to-speech (TTS) or speech-language modeling pipelines, especially under large vocabulary sizes (e.g., 65,536 tokens) and shorter temporal sequences. Analyzing token-level perplexity, attention behavior, and reconstruction fidelity in LLM-based TTS finetuning could help determine whether lower frame rates hinder or help generative quality.

\section{Acknowledgement}

Special thanks to the people at home for their patience and for not getting too upset about the electricity bill caused by running two RTX 3090 Ti GPUs around the clock for 45 days. I am gratefully acknowledge my manager for pushing me to publish and for their continued encouragement.

\bibliographystyle{plainnat}
\bibliography{neurips_2023}

@misc{ye2024codecdoesmatterexploring,
      title={Codec Does Matter: Exploring the Semantic Shortcoming of Codec for Audio Language Model}, 
      author={Zhen Ye and Peiwen Sun and Jiahe Lei and Hongzhan Lin and Xu Tan and Zheqi Dai and Qiuqiang Kong and Jianyi Chen and Jiahao Pan and Qifeng Liu and Yike Guo and Wei Xue},
      year={2024},
      eprint={2408.17175},
      archivePrefix={arXiv},
      primaryClass={eess.AS},
      url={https://arxiv.org/abs/2408.17175}, 
}

@misc{ardila2020commonvoicemassivelymultilingualspeech,
      title={Common Voice: A Massively-Multilingual Speech Corpus}, 
      author={Rosana Ardila and Megan Branson and Kelly Davis and Michael Henretty and Michael Kohler and Josh Meyer and Reuben Morais and Lindsay Saunders and Francis M. Tyers and Gregor Weber},
      year={2020},
      eprint={1912.06670},
      archivePrefix={arXiv},
      primaryClass={cs.CL},
      url={https://arxiv.org/abs/1912.06670}, 
}

@misc{baba2024t05voicemoschallenge2024,
      title={The T05 System for The VoiceMOS Challenge 2024: Transfer Learning from Deep Image Classifier to Naturalness MOS Prediction of High-Quality Synthetic Speech}, 
      author={Kaito Baba and Wataru Nakata and Yuki Saito and Hiroshi Saruwatari},
      year={2024},
      eprint={2409.09305},
      archivePrefix={arXiv},
      primaryClass={cs.SD},
      url={https://arxiv.org/abs/2409.09305}, 
}

@misc{hsu2021hubertselfsupervisedspeechrepresentation,
      title={HuBERT: Self-Supervised Speech Representation Learning by Masked Prediction of Hidden Units}, 
      author={Wei-Ning Hsu and Benjamin Bolte and Yao-Hung Hubert Tsai and Kushal Lakhotia and Ruslan Salakhutdinov and Abdelrahman Mohamed},
      year={2021},
      eprint={2106.07447},
      archivePrefix={arXiv},
      primaryClass={cs.CL},
      url={https://arxiv.org/abs/2106.07447}, 
}

@inproceedings{siuzdak2024snac,
  title={SNAC: Multi-Scale Neural Audio Codec},
  author={Siuzdak, Hubert and Gr{\"o}tschla, Florian and Lanzend{\"o}rfer, Luca A},
  booktitle={Audio Imagination: NeurIPS 2024 Workshop AI-Driven Speech, Music, and Sound Generation},
  year={2024}
}

@misc{defossez2022highfidelityneuralaudio,
      title={High Fidelity Neural Audio Compression}, 
      author={Alexandre Défossez and Jade Copet and Gabriel Synnaeve and Yossi Adi},
      year={2022},
      eprint={2210.13438},
      archivePrefix={arXiv},
      primaryClass={eess.AS},
      url={https://arxiv.org/abs/2210.13438}, 
}

@misc{kumar2023highfidelityaudiocompressionimproved,
      title={High-Fidelity Audio Compression with Improved RVQGAN}, 
      author={Rithesh Kumar and Prem Seetharaman and Alejandro Luebs and Ishaan Kumar and Kundan Kumar},
      year={2023},
      eprint={2306.06546},
      archivePrefix={arXiv},
      primaryClass={cs.SD},
      url={https://arxiv.org/abs/2306.06546}, 
}

@misc{defossez2024moshispeechtextfoundationmodel,
      title={Moshi: a speech-text foundation model for real-time dialogue}, 
      author={Alexandre Défossez and Laurent Mazaré and Manu Orsini and Amélie Royer and Patrick Pérez and Hervé Jégou and Edouard Grave and Neil Zeghidour},
      year={2024},
      eprint={2410.00037},
      archivePrefix={arXiv},
      primaryClass={eess.AS},
      url={https://arxiv.org/abs/2410.00037}, 
}

@misc{zhang2024speechtokenizerunifiedspeechtokenizer,
      title={SpeechTokenizer: Unified Speech Tokenizer for Speech Large Language Models}, 
      author={Xin Zhang and Dong Zhang and Shimin Li and Yaqian Zhou and Xipeng Qiu},
      year={2024},
      eprint={2308.16692},
      archivePrefix={arXiv},
      primaryClass={cs.CL},
      url={https://arxiv.org/abs/2308.16692}, 
}

@misc{ji2025wavtokenizerefficientacousticdiscrete,
      title={WavTokenizer: an Efficient Acoustic Discrete Codec Tokenizer for Audio Language Modeling}, 
      author={Shengpeng Ji and Ziyue Jiang and Wen Wang and Yifu Chen and Minghui Fang and Jialong Zuo and Qian Yang and Xize Cheng and Zehan Wang and Ruiqi Li and Ziang Zhang and Xiaoda Yang and Rongjie Huang and Yidi Jiang and Qian Chen and Siqi Zheng and Zhou Zhao},
      year={2025},
      eprint={2408.16532},
      archivePrefix={arXiv},
      primaryClass={eess.AS},
      url={https://arxiv.org/abs/2408.16532}, 
}

@misc{jiang2025unicodecunifiedaudiocodec,
      title={UniCodec: Unified Audio Codec with Single Domain-Adaptive Codebook}, 
      author={Yidi Jiang and Qian Chen and Shengpeng Ji and Yu Xi and Wen Wang and Chong Zhang and Xianghu Yue and ShiLiang Zhang and Haizhou Li},
      year={2025},
      eprint={2502.20067},
      archivePrefix={arXiv},
      primaryClass={eess.AS},
      url={https://arxiv.org/abs/2502.20067}, 
}

@misc{wang2025unittsendtoendttsdecoupling,
      title={UniTTS: An end-to-end TTS system without decoupling of acoustic and semantic information}, 
      author={Rui Wang and Qianguo Sun and Tianrong Chen and Zhiyun Zeng and Junlong Wu and Jiaxing Zhang},
      year={2025},
      eprint={2505.17426},
      archivePrefix={arXiv},
      primaryClass={cs.SD},
      url={https://arxiv.org/abs/2505.17426}, 
}

@misc{julian2025finitescalarquantizationenables,
      title={Finite Scalar Quantization Enables Redundant and Transmission-Robust Neural Audio Compression at Low Bit-rates}, 
      author={Harry Julian and Rachel Beeson and Lohith Konathala and Johanna Ulin and Jiameng Gao},
      year={2025},
      eprint={2509.09550},
      archivePrefix={arXiv},
      primaryClass={cs.SD},
      url={https://arxiv.org/abs/2509.09550}, 
}

@misc{zhao2025longcataudiocodecaudiotokenizerdetokenizer,
      title={LongCat-Audio-Codec: An Audio Tokenizer and Detokenizer Solution Designed for Speech Large Language Models}, 
      author={Xiaohan Zhao and Hongyu Xiang and Shengze Ye and Song Li and Zhengkun Tian and Guanyu Chen and Ke Ding and Guanglu Wan},
      year={2025},
      eprint={2510.15227},
      archivePrefix={arXiv},
      primaryClass={eess.AS},
      url={https://arxiv.org/abs/2510.15227}, 
}

@misc{hu2026qwen3ttstechnicalreport,
      title={Qwen3-TTS Technical Report}, 
      author={Hangrui Hu and Xinfa Zhu and Ting He and Dake Guo and Bin Zhang and Xiong Wang and Zhifang Guo and Ziyue Jiang and Hongkun Hao and Zishan Guo and Xinyu Zhang and Pei Zhang and Baosong Yang and Jin Xu and Jingren Zhou and Junyang Lin},
      year={2026},
      eprint={2601.15621},
      archivePrefix={arXiv},
      primaryClass={cs.SD},
      url={https://arxiv.org/abs/2601.15621}, 
}

\end{document}